\documentclass[letterpaper, 10 pt, journal, twoside]{IEEEtran}

\pdfoutput=1

\usepackage{cite}
\usepackage{graphicx}
\usepackage{amsmath} 
\usepackage{amssymb}  
\usepackage{myMacros}
\usepackage[caption=false,font=footnotesize]{subfig}
\usepackage[noend]{algorithmic}
\usepackage{lipsum}

\usepackage[dvipsnames]{xcolor}

\begin{document}
\title{Data-efficient Model Learning and Prediction for Contact-rich Manipulation Tasks}

\author{Shahbaz A. Khader, Hang Yin, Pietro Falco and Danica Kragic,~\IEEEmembership{Fellow,~IEEE}%
\thanks{Manuscript received: February, 11, 2020; Accepted April, 29, 2020.}
\thanks{This paper was recommended for publication by Editor Dongheui Lee upon evaluation of the Associate Editor and Reviewers' comments.
This work was partially supported by the Wallenberg AI, Autonomous Systems and Software Program (WASP) funded by the Knut and Alice Wallenberg Foundation. Corresponding author: Shahbaz A. Khader} 
\thanks{Shahbaz A. Khader is with RPL, EECS, KTH, Stockholm, Sweden and ABB Future Labs, Baden, Switzerland
        {\tt\footnotesize shahak@kth.se}.}%
\thanks{Hang Yin and Danica Kragic are with RPL, EECS, KTH, Stockholm, Sweden
        {\tt\footnotesize \{hyin, dani\}@kth.se}}%
\thanks{Pietro Falco is with ABB Corporate Research, Vasteras, Sweden
        {\tt\footnotesize pietro.falco@se.abb.com}}%
\thanks{Digital Object Identifier (DOI): see top of this page.}
}


\markboth{IEEE Robotics and Automation Letters. Preprint Version. Accepted April, 2020}
{Khader \MakeLowercase{\textit{et al.}}: Data-efficient Model Learning and Prediction for Contact-rich Manipulation Tasks}

\maketitle

\begin{abstract}
In this paper, we investigate learning forward dynamics models and multi-step prediction of state variables (long-term prediction) for contact-rich manipulation. The problems are formulated in the context of model-based reinforcement learning (MBRL). We focus on two aspects--discontinuous dynamics and data-efficiency--both of which are important in the identified scope and pose significant challenges to state-of-the-art methods. We contribute to closing this gap by proposing a method that explicitly adopts a specific hybrid structure for the model while leveraging the uncertainty representation and data-efficiency of Gaussian process. Our experiments on an illustrative moving block task and a 7-DOF robot demonstrate a clear advantage when compared to popular baselines in low data regimes.
\end{abstract}

\begin{IEEEkeywords}
Model Learning for Control, Contact Modeling, Reinforcement Learning
\end{IEEEkeywords}

%
\IEEEpeerreviewmaketitle

\section{Introduction}\label{sec:intro}
\IEEEPARstart{M}{odels} of forward dynamics have been extensively used for planning and control problems such as model-based policy search \cite{deisenroth2013survey} and model predictive control (MPC). By sequencing multiple one-step predictions an arbitrary number of future predictions can be obtained, a process we refer to as long-term prediction. Long-term prediction offers a simulated rollout of a dynamical system, such as a manipulator, that can be used to evaluate a control policy under delayed rewards. Model-based reinforcement learning (MBRL) methods are formed by combining this idea with a data-driven paradigm of learning the model itself. An interesting application for MBRL is contact-rich manipulation, where it is notoriously difficult to obtain analytical models of the highly complex interaction dynamics. 

Learning to predict in a contact-rich environment poses some unique challenges to model learning approaches. First, establishing contact under a rigid assumption, which is common in an industrial setting, is often modeled as discontinuous dynamics with an impulsive velocity update \cite{Bender06}. This is often overlooked by existing learning models, possibly leading to an inferior performance. Secondly, applying model learning in MBRL is more demanding in terms of data-efficiency because a high number of trials could cause wear and tear. Although this paper focuses on model learning and long-term prediction for contact tasks, we are constrained by the larger MBRL consideration of data-efficiency.

A general concern for any model learning scenario is prediction uncertainties. Because of the recursive nature of long-term prediction, even a small error in the model can cause a significant divergence from the real system behavior. Furthermore, erroneous predictions can follow if the system moves into a region that has seen less training data \cite{ross2011reduction}, something which is only exaggerated by the requirement of learning from sparse and discontinuous data. An effective strategy, as introduced in \cite{deisenroth2015gaussian}, is to adopt a probabilistic approach that would directly encode the uncertainty due to lack of training data in the long-term prediction and then utilize this representation in policy optimization. 

A number of recent MBRL methods have been successfully demonstrated in some contact-involving tasks such as manipulation \cite{lenz2015deepmpc}, \cite{levine2015learning} and locomotion \cite{nagabandi2018neural}, \cite{chua2018deep}. Since these methods focused only on the overall MBRL performance, questions remain about the most effective strategy to achieve model learning and long-term prediction. In answering this question, we adopt a strategy of hybrid dynamical system~\cite{lunze2009handbook} that directly addresses the discontinuous nature of dynamics and compares it with highly flexible state-of-the-art methods. Specifically, we propose to learn a model with a structure resembling a \textit{hybrid automata}~\cite{lunze2009handbook}, in which Gaussian process (GP) is used to model the subsystems (or modes). We proceed further and provide a solution for probabilistic long-term prediction using such a hybrid model. Results from a contact motion experiment performed on an industrial manipulator clearly show that the proposed method outperforms the baselines by a large margin. To the best of our knowledge, such a direct evaluation of long-term prediction of contact-rich manipulation tasks has not been undertaken before.

\section{Related Work}\label{sec:rw}
Although model learning for robot control has been actively researched \cite{nguyen2011model}, learning forward dynamics models for contact-rich manipulation tasks that feature discontinuities has been less explored.  Recent works that target contact-rich tasks, such as  \cite{calandra2016manifold}\cite{levine2015learning}\cite{lenz2015deepmpc}, do not include a direct study on the performance of probabilistic long-term prediction.

A prominent example of probabilistic model learning for MBRL is PILCO \cite{deisenroth2015gaussian}, in which GPs were used for model learning. Unfortunately, with squared exponential (SE) kernels, GP does not support learning and prediction through discontinuous dynamics. A strategy for addressing discontinuities within the GP framework is manifold GP \cite{calandra2016manifold}. The SE kernel function operates in a feature space that is transformed from the input space using an ANN. However, the method was evaluated only for single step prediction. A more recent work (PETS) \cite{chua2018deep} that learns an uncertainty aware ANN model succeeded in regaining many of the desirable features of GP while retaining the expressive power and scalability of ANN. However its applicability for contact-rich manipulation in low data regime is not well established.


Also relevant to our work is model learning with multistep prediction for contact-rich tasks such as \cite{nagabandi2018neural} and \cite{lenz2015deepmpc}, in which deterministic dynamics models were learned using ANN. An uncertainty-aware approach, such as ours, can potentially help achieve greater data efficiency for MBRL \cite{deisenroth2015gaussian}. Finally, Levine et al. \cite{levine2015learning} used Gaussian mixture models (GMM) to learn locally linear model priors. While each component of the GMM represents a locally linear dynamics mode, our method benefits from a more flexible nonlinear GP model.

Hybrid systems approach has been considered for robotics planning in the past \cite{posa2013direct}. Most learning applications consider a switched systems form, in which a number of local models (modes) and a selector function based on input is learned \cite{paoletti2007identification} \cite{lauer2014piecewise}. This is related to the more general mixture of experts idea \cite{jacobs1991adaptive}. A GP based switched system for contact tasks is proposed in \cite{calandra2015learninga}. Improvement on the switched system is achieved by allowing mode predictions that consider the current mode as well \cite{linderman2017bayesian} \cite{santana2015learning}. None of these models, however, can support discontinuous state evolution in long-term prediction.

In all the cases mentioned so far, complex models can indeed be learned but when used recursively for long-term prediction they are not capable of generating a discontinuous long-term prediction (see Sec. \ref{sec:prob_form}). Our method addresses this by directly targeting the general structure of hybrid automata \cite{lunze2009handbook} that includes a concept called reset maps (see Sec. \ref{sec:ha}) that exclusively handles the issue of discontinuous state. Furthermore, we simplify the learning process with the assumption of linear separability (in feature space) of modes (see Sec. \ref{sec:clustering}) and follow it with a solution for long-term prediction. A previous attempt that shares some similarities with our method is presented in \cite{lee2017unsupervised}. But, the suggested approach, that involves iterative GP training, would hardly scale for realistic cases of robotic manipulation. The authors did not validate their method with any robotics experiments.

\begin{figure}
\centering
  \includegraphics[width=0.45\textwidth]{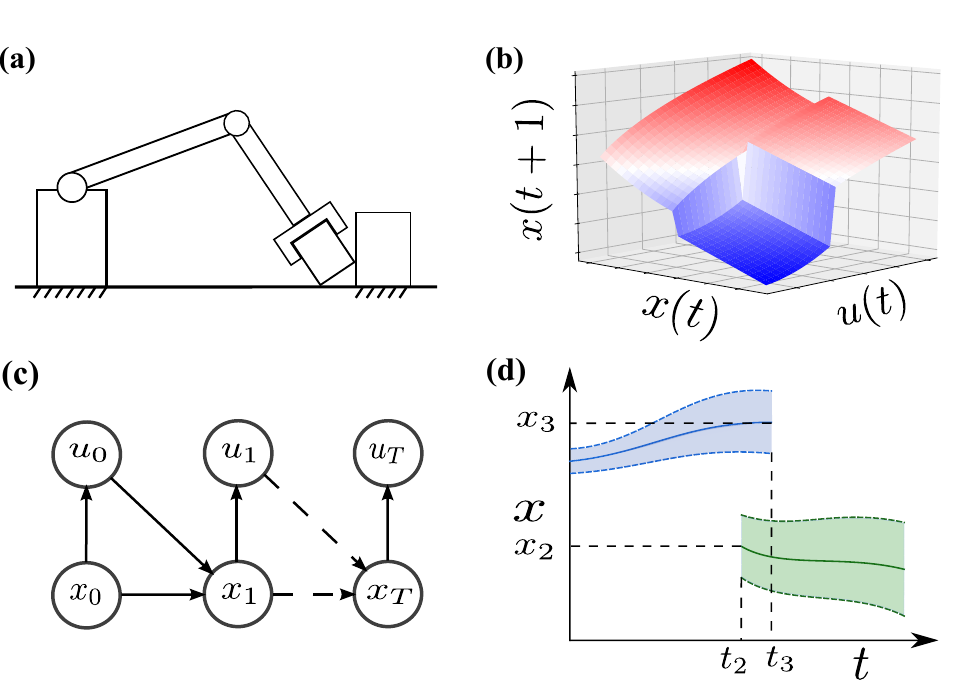}
    \caption{(\textbf{a}) Contact with static and rigid objects (illustrative) (\textbf{b}) Piecewise smooth dynamics (\textbf{c}) Long-term prediction (\textbf{d}) Discontinuous trajectory distribution: trajectory is reinitialized from $x_3$ to $x_2$ after mode change. The state $x$ can be thought of as velocity.}
  \label{fig:prob_fig}
\end{figure}

\section{Background and Problem Formulation}\label{sec:prob_form}
\subsection{Model Learning and Long-term Prediction}\label{sec:general_model_learning}
The goal of model learning is, given a training dataset consisting of tuples $(\vxi, \vui)$, to fit a model of the form,  
\begin{align}\label{eqn:dynamics_gen}
   p(\vxi[t+1]|\vxi[t], \vui[t]) &= \mathcal{N}(\mu(\vxi[t], \vui[t]), \Sigma(\vxi[t], \vui[t])),
\end{align}
where $\vxi=\begin{bmatrix} \vqi^T & \vqdi^T \end{bmatrix}^T$ is the state vector comprising of the joint positions $\vqi\in\mathbb{R}^{D}$ and velocities $\vqdi\in\mathbb{R}^{D}$, $\vui\in\mathbb{R}^{D}$ is the control action taken by a policy, $D$ is the number of axes of the manipulator, $\mu(.)$ and $\Sigma(.)$ are the mean and variance functions of the dynamics model. The main challenge in model learning for contact-rich manipulation is the piecewise continuous nature of dynamics (Fig. 1b). Each of the smooth regions govern motions in a particular mode of operation (or modes) such as free motion, sliding against kinematic constraints, or blocked conditions. The discontinuities correspond to an abrupt change from one mode to another. 

In long-term prediction (Fig. 1c), we seek the prediction of the state distributions $p(\vxi[1]|\pi), p(\vxi[2]|\pi), ..., p(\vxi[T]|\pi)$, starting from an initial state distribution $p(\vxi[0])$, for the purpose of evaluating a policy $\pi = p(\vui|\vxi)$. $T$ is the number of time steps assumed for the task. The propagation of uncertainties from one time step to the next is not trivial for a nonlinear model and we refer to \cite{deisenroth2015gaussian} for more details. A unique feature of robotic systems is that when they collide with the environment, in addition to a mode change, there can also be an abrupt change in velocity due to loss of energy (Fig. 1d). In addition to this discontinuity, the trajectory distribution can exhibit multimodality either due to uncertainty in the exact time of impact ($t_2<t_3$ in Fig. 1d) or alternate possibilities post contact (e.g. stick or slip). It is extremely difficult for any standard method to learn and generate such long-term prediction sequences from little data.

\begin{figure*}[t]
  \includegraphics[width=\textwidth]{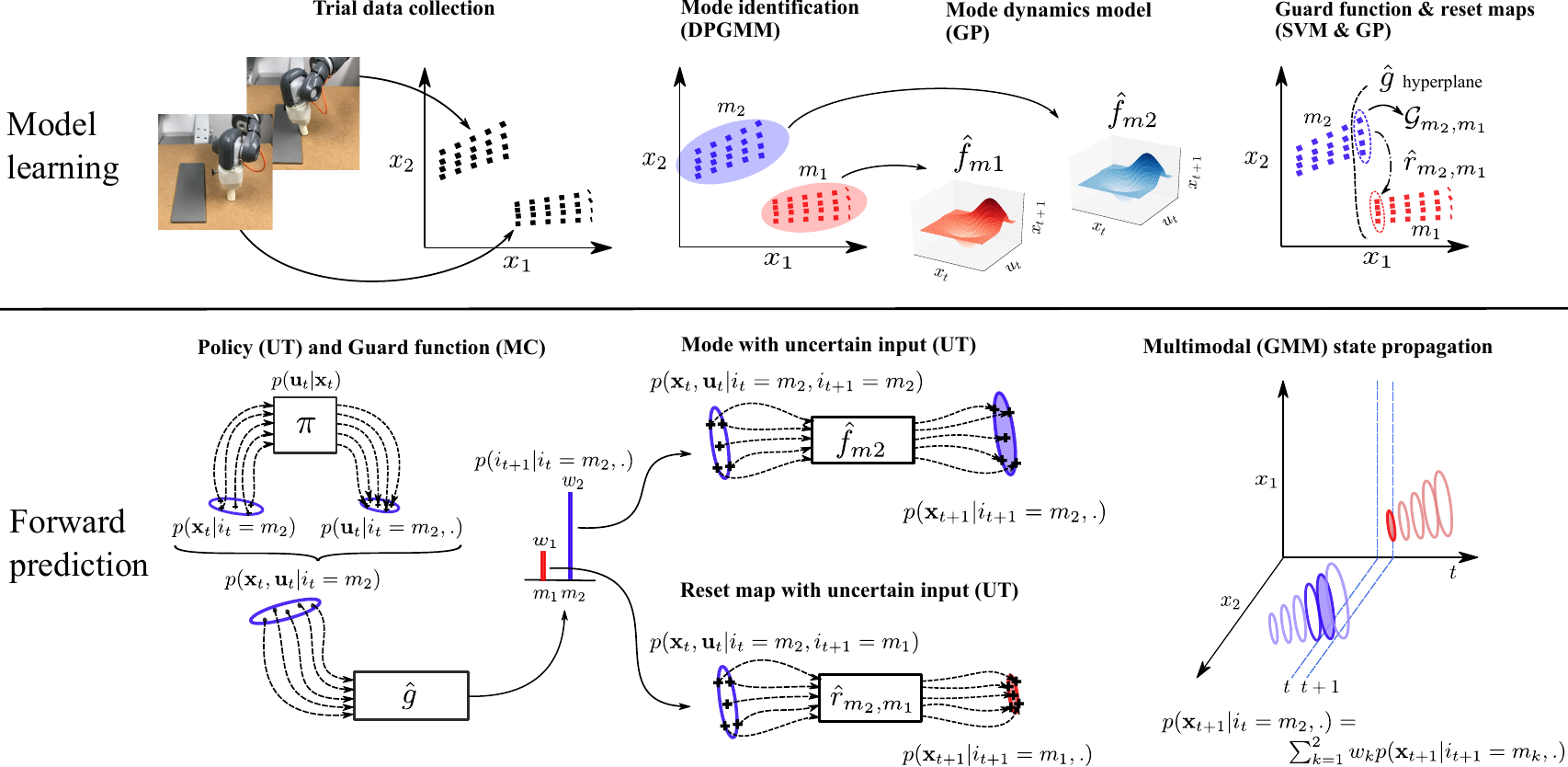}
  \caption{Model-learning and forward prediction through a model with the structure of a hybrid automaton: (Top) Training models for modes ($\hat{f}_i$), guard function ($\hat{g}$) and reset maps ($\hat{r}_{i,j}$). (Bottom) Unscented transform (UT) based uncertainty propagation through policy ($\pi$), $\hat{f}_i$, and $\hat{r}_{i,j}$; Monte Carlo (MC) based uncertainty propagation through $\hat{g}$. Splitting of trajectory distribution is also illustrated.}
  \label{fig:our_meth}
  \vspace{-1em}
\end{figure*}

\subsection{Hybrid System View of Dynamics}\label{sec:ha}
Hybrid Automata (HA) \cite{lunze2009handbook} can be used to model systems that consist of both continuous and discrete states. Let $i_t\in\mathcal{M}$ represent the mode at time $t$ where $\mathcal{M}$ is a set of discrete modes. In a given mode $i_t$, the continuous state variable $\vxi\in\mathcal{X}_i$ is controlled by the control action $\vui \in \mathcal{U}$ subject to its own dynamics $f_i$. $\mathcal{X}_i$ is the region in state space where $f_i$ is defined and $\mathcal{U}$ is the action space of the system.

The main elements that define a hybrid automaton are (for a more complete definition see \cite{lunze2009handbook}): 
\begin{align}
    f_i&: \mathcal{X}_i\times\mathcal{U}\rightarrow\mathcal{X}_i, \quad i\in\mathcal{M}\textrm{ (mode dynamics)}\label{eqn:ha_mode}\\
    \mathcal{T}&\subseteq\mathcal{M}\times\mathcal{M}\qquad \qquad \qquad  \textrm{(transition relations)}\label{eqn:ha_T}\\
    \mathcal{G}_{i,j}&\subset\mathcal{X}_i\times \mathcal{U}, \quad (i,j)\in\mathcal{T}\qquad \textrm{(guard region)}\label{eqn:ha_guard}\\
    \mathcal{R}_{i,j}&:\mathcal{X}_i\times\mathcal{U}\rightarrow \mathcal{X}_{j}, \quad (i,j)\in\mathcal{T}\quad \textrm{(reset map)}\label{eqn:ha_rmap}
\end{align}
$\mathcal{T}$ represents the allowed mode transitions, $\mathcal{G}_{i,j}$ is the state-action space region that triggers the transition $i\rightarrow j$ and $\mathcal{R}_{i,j}$ is a function that maps ($\vxi[t]^-,\vui[t]^-$) (before transition) to $\vxi[t+1]^+$ (after transition) for the transition $i\rightarrow j$. Now it can be seen that a hybrid state ($\vxi, i_t$) can be used to represent the overall system state, and its evolution, under the influence of the control input $\vui$ and a given initial state ($i_0, \vxi[0]$), can represent evolution of piecewise nonlinear dynamics such as contact-rich manipulation.

\subsection{Hybrid Models in MBRL} \label{sec:hybrid_models_in_mbrl}

When considering one-step prediction of state  $\vxi[t+1]$ in MBRL, one source of uncertainty is the inherent stochasticity in the system (system noise). Another is the propagated uncertainty of the previous state $\vxi[t]$. Lastly, an important source, in the context of MBRL, is the uncertainty due to lack of training data. The last case, if represented, can be exploited to address the issue of state distribution mismatch during training and prediction, also called the DAgger \cite{ross2011reduction} effect. Even for a hybrid system of the form (\ref{eqn:ha_mode}-\ref{eqn:ha_rmap}), a general solution has to account for the three independent sources of uncertainties whenever possible. In this paper, we do not consider observation noise.

We consider the dynamics of a manipulator interacting with an environment. In an MBRL (or any RL) context, where the action space is joint torque, the definition of environment includes the manipulator as well. For the state of such an environment to be adequately represented by the joint positions and velocities, i.e, preserving the Markov property, it is necessary for all objects in the environment (including the manipulator) to be rigid and all objects excluding the manipulator to be static. Fortunately, this requirement is not too limiting and covers many industrial applications such as assembly. Under this assumption the regions of state space corresponding to each mode do not intersect ($\mathcal{X}_i\cap \mathcal{X}_j=\emptyset$ for $i\ne j)$.

\section{Model Learning and Long-term Prediction}\label{sec:the_method}
Our method touches upon all elements in (\ref{eqn:ha_mode}-\ref{eqn:ha_rmap}) and has two main aspects: model learning (Algorithm 1) and uncertainty propagation (Algorithm 2). They are also visualized in Fig. \ref{fig:our_meth}. It involves learning dynamics modes $f_i$, a transition relation $\mathcal{T}$, a guard function that act on the guard regions $\mathcal{G}_{i,j}$ and reset maps $\mathcal{R}_{i,j}$, for $i\in\mathcal{M}$ and $(i,j)\in\mathcal{T}$. After presenting mode learning, with some components being probabilistic and others not, we focus on the forward propagation of input uncertainties. Finally, multimodal behavior in long-term prediction is tackled.

\subsection{Learning Models with Hybrid Structure}
\subsubsection{Mode discovery through clustering}\label{sec:clustering}
We observe that in addition to $\mathcal{X}_i\cap \mathcal{X}_j=\emptyset$ for $i\ne j$ (Sec. \ref{sec:hybrid_models_in_mbrl}), the data from a set of MBRL rollouts will feature well separated clusters in state space due to discontinuous change in velocities during contact. This is exploited for mode discovery by clustering the trial data ($\mathbb{D}^{\vec{x}}$) in state space or an appropriate feature space. For our case we propose a feature vector defined as concatenation of position and velocities of three non-collinear points fixed at the end-effector and expressed in Cartesian coordinates. The motivation being that impacts from contact are more pronounced at the end-effector and three points can represent both position and orientation. We use the Dirichlet Process Gaussian Mixture Model (DPGMM) \cite{blei2006variational} clustering method that automatically infers the number of clusters (modes) $K$. Fig. \ref{fig:our_meth} symbolically depicts the clustering in a two-dimensional state space. We maintain the sequential order of the dataset and the cluster labels $\mathbf{z}$ for later use.

\begin{figure}[t]
\hrule
\vspace{0.2em}
 \textbf{Algorithm 1}\quad Model learning for hybrid structure\label{alg:model_learn}
  \vspace{0.2em}
  \hrule
  \begin{algorithmic}[1]
  \vspace{0.2em}
  \REQUIRE Temporally ordered data set $\mathbb{D}$
  \STATE $\vec{z},K\gets$ DPGMM cluster on $\mathbb{D}^{\vec{x}}$\qquad(cluster on state data)\label{alg_ln:model_learn_clust}
   \FOR{each cluster $i=1,...,K$}
   		\STATE{$\hat{f}_{i}(\vxi, \vui)\gets
   		GPR(\mathbb{D}_i^{\vec{x}\vec{u}},\mathbb{D}_i^{\Delta\mathbf{x}})$}\quad(Mode dynamics)
   \ENDFOR
   \FOR{each mode switch $i\rightarrow j$}
   		\STATE{$\hat{r}_{i,j}(\vxi, \vui)\gets 
   		GPR(\mathbb{D}_{ij}^{\vec{x}\vec{u}},\mathbb{D}_{ij}^{\vec{x'}})$}\quad(Reset map)
   \ENDFOR
   \STATE{$\hat{g}(\vxi, \vui)\gets SVM\_train(\mathbb{D}^{\vec{x}\vec{u}},\vec{z}')$}\quad(Guard function)
  \end{algorithmic}
  \vspace{0.2em}
  \hrule
\end{figure}


\subsubsection{Dynamics Modes}
Individual dynamics modes $f_i$ represented in (\ref{eqn:ha_mode}) have to be learned from data. In order to maximize data efficiency and to encode uncertainty due to lack of training data (also the system noise) we chose Gaussian process regression (GPR) as the preferred method. The effectiveness of GPs for MBRL has been demonstrated in \cite{deisenroth2015gaussian}. The difference in our case being that it is in a hybrid system context. Following \cite{deisenroth2015gaussian}, we learn the form in (\ref{eqn_ln:dyn_noswitch_delta}), but to maintain simplicity the notation $\hat{f}_i$ (learned model) assumes the form (\ref{eqn:dynamics_gen}).
\begin{align}
    p(\Delta\vxi|i_t, \vxi[t], \vui[t]),\ \vxi[t+1] = \vxi+\Delta\vxi \label{eqn_ln:dyn_noswitch_delta}
\end{align}
The conditioning on $i_t$ in (\ref{eqn_ln:dyn_noswitch_delta}) is to indicate that $i_t$ influences $\Delta\vxi$ by selecting the correct $f_i$. We train a multioutput GP dynamics model (independent outputs) for each cluster with its respective data $\{\mathbb{D}_i^{\vec{xu}}, \mathbb{D}_i^{\Delta \vec{x}}\}$.

A GP is a distribution of functions $f\sim\mathcal{GP}(m,k)$ completely defined by a mean function $m$ and covariance function $k$. In our case, $m\equiv0$ and $k$ is the standard SE function with automatic relevance determination (ARD). In our setting, we obtain independent uncertainty estimation for each output dimension of each mode. The models are trained by optimizing the log marginal likelihood.

\subsubsection{Transition Relations}
The transition relations in (\ref{eqn:ha_T}) represent the allowed transitions of the model. This information can be easily extracted from the labelled data (post-clustering) in which the sequential order was preserved. All mode transitions present in the sequential data are considered as allowed and stored in the transition relation $\mathcal{T}$. We model a deterministic $\mathcal{T}$ because the guard function that actually triggers transitions is deterministic (see below). Generalization to a new unseen mode is discussed in Sec. \ref{sec:disscuss}.

\subsubsection{Guard functions}
The guard regions $\mathcal{G}_{i,j}$ (\ref{eqn:ha_guard}) specify regions in the state-action space which if entered will result in the mode transition $i\rightarrow j$. The guard concept is practically implemented as a guard function $g(.)$ that predicts the next mode $i_{t+1}$ as a function of ($\vxi$, $\vui$).
\begin{align}
    i_{t+1} = g(\vxi, \vui)
    \label{eqn_ln:mode_pred}
\end{align}
It could be a function of $i_t$ also \cite{santana2015learning}, but we assume that such an extra layer of dynamics is unnecessary for our case. If the predicted mode $i_{t+1}$ is not the same as the current mode $i_t$, which is always known, then a transition must follow provided the transition is valid in $\mathcal{T}$. We assume that $\mathcal{G}_{i,j}\cap\mathcal{G}_{i,k}=\emptyset$ for $j\ne k$, which implies that the guard function is deterministic. 

Learning $g(.)$ is a classification problem and consequently we use a multiclass support vector machine (SVM) classification to obtain an approximation $\hat{g}(.)$. It is trained using the dataset $\{\mathbb{D}^{\vec{xu}},\vec{z}'\}$, where  $\mathbb{D}^{\vec{xu}}$ and $\vec{z}'$ consists of arrays of $(\vxi, \vui)$ and  $z_{t+1}$, respectively. The SVM hyperplane (in $\vec{x}$-space) is symbolically depicted in the upper right part of Fig. \ref{fig:our_meth}. This is another use of the sequential order of the dataset.

\subsubsection{Reset Maps}
Reset maps play a very important role in our method. Without it discontinuous state evolution will not be possible. The purpose of a reset map is to map a guard region $\mathcal{G}_{i,j}$ to an appropriate region in $\mathcal{X}_{j}$ for the transition $i\rightarrow j$. For the same reasons given for mode learning, we use GPR for learning probabilistic reset maps $r_{i,j}(\vxi,\vui)$, which being dependent on $i_t$ and $i_{t+1}$, can also be expressed as:
\begin{align}
    p(\vxi[t+1]|i_t, i_{t+1}, \vxi[t], \vui[t]),\ i_t\ne i_{t+1} \label{eqn_ln:dyn_switch}
\end{align}
For every $(i,j)\in\mathcal{T}$, a multioutput GP $\hat{r}_{i,j}(.)$ is learned to approximate $r_{i,j}(.)$. The training data, $\{\mathbb{D}_{ij}^{\vec{xu}}, \mathbb{D}_{ij}^{\vec{x'}}\}$, for each reset map is appropriately extracted from the clustered dataset that had the sequential order of data preserved.

\subsection{Uncertainty Propagation through the Learned Model}\label{sec:uncert_prop}
A single step in long-term prediction involves propagation of input uncertainties through $\hat{f}_i(.)$, $\hat{g}(.)$ and $\hat{r}_{i,j}(.)$. The input uncertainties originate from $p(\vxi[0])$ and the stochastic policy $\pi$ and exist even if all models are deterministic, which in our case is only $\hat{g}(.)$.

\subsubsection{Dynamics mode}
Both the policy and the mode dynamics have its own intrinsic uncertainty which gets combined with the propagated input uncertainties. We use a particle based method \cite{ko2009gp} that is based on the unscented transform (UT) method \cite{julier1997new} for this. An analytic but less computationally efficient alternative for GP is moment matching \cite{deisenroth2009analytic}.

If the UT method is represented by the operator $UT(f(x),p(x))$, where $f(x)$ is any nonlinear function through which a distribution $p(x)$ has to be propagated, the one-step forward propagation through a mode $\hat{f}_i(.)$ is summarized as: 
\begin{subequations}\label{eqn:one_step_mode_prop}
\begin{align}
    p(\vui|\vxi,i_t) &\leftarrow UT(\pi, p(\vxi|i_t))\label{eqn_ln:pol_pred_1_step}\\
    p(\vxi, \vui|i_t) &\leftarrow p(\vxi|i_t)p(\vui|\vxi,i_t)\label{eqn_ln:p_xu_1_step}\\
    p(\vxi[t+1]|i_t,i_{t+1},.) &\leftarrow UT(\hat{f}_{i_t},p(\vxi, \vui|i_t,i_{t+1}))\label{eqn_ln:nosw_1_step}
\end{align}
\end{subequations}
where $i_{t+1}$ is predicted to be equal to $i_t$ (no transition). Bottom half of Fig. \ref{fig:our_meth} shows the process with the UT particles indicated as crosses.

\subsubsection{Guard Function}\label{sec:guard_pred}
We achieve forward propagation for the deterministic $\hat{g}(.)$ by using a simple Monte Carlo (MC) approach. A number of samples are drawn from the input distribution $p(\vxi, \vui)$ and the relative frequencies of the predicted classes are approximated as a discrete probability distribution $p(i_{t+1})$ (MC particles represented as dots in Fig. \ref{fig:our_meth}). If the MC process is represented with the operator $MC(f(x),p(x))$, where $f(x)$ and $p(x)$ are the predictor and the input distribution, respectively, we have:
\begin{align}
    p(i_{t+1}|i_t,.) &\leftarrow MC(\hat{g}(.),p(\vxi,\vui|i_t))\label{eqn_ln:mode_pred_1_step}
\end{align}

\subsubsection{Reset Map}\label{sec:rm_pred}
Following the same approach as in the case of mode dynamics, we use the UT method for input uncertainty propagation.
\begin{align}
    p(\vxi[t+1]|i_t,i_{t+1},.) &\leftarrow  UT(\hat{r}_{i_t,i_{t+1}},p(\vxi,\vui|i_t,i_{t+1}))\label{eqn_ln:sw_1_step}
\end{align}
Here $i_{t+1}$ is predicted to be not equal to $i_t$ (a transition) and $(i_t,i_{t+1})\in\mathcal{T}$. The reset map $\hat{r}_{i,j}$ has been trained with input data that belongs to the guard $\mathcal{G}_{i,j}$. During forward propagation, only a distribution $p(\vxi,\vui)$ that is strictly in $\mathcal{G}_{i,j}$ should be propagated through the corresponding $\hat{r}_{i,j}$. The Monte Carlo approach used for $\hat{g}(.)$ provides an easy solution by allowing only those particles that triggered the guard to be used to approximate an appropriate $p(\vxi,\vui)$.

\subsection{Probabilistic Switching and Multimodality in Long-term prediction}\label{sec:long_pred_form}

The probabilistic nature of state evolution implies that the exact time of a mode transition is not deterministic ($t_2<t_3$ in Fig. 1d). Staying with Fig. 1d, the guard function will successively predict nonzero probabilities for both modes at every time step from $t_2$ to $t_3$. The reset map of the blue mode will also make successive predictions into the region of the green mode (around $x_2$) at these time steps. All the incoming predictions at the green mode cannot be propagated in parallel but has to be merged to maintain a unimodal Gaussian state. Meanwhile, the blue mode propagates forward the remaining state distribution after the successive splitting process mentioned in Sec. \ref{sec:rm_pred}. Sometimes the original mode may continue and then we will have a multimodal (blue and green modes, for example) state evolution for longer periods. This can happen when such a trend exists in the training data, for example in the case of a stochastic stick-slip phenomenon. 

The switching process is called probabilistic switching (Algorithm 2). We denote a smooth evolving prediction sequence within a mode as a segment $s$. A transition from $s$ to $s'$ is represented as $s|s'$ and probability weights as $w$. All subscripts are time indices and superscripts are associations to $s$ or $s|s'$. We distinguish between $s$ and $i$ because multiple segments may be formed for the same $i$. Line \ref{alg_ln:split} represents the process of splitting state-action distributions. The merging of distributions are indicated in lines \ref{alg_ln:update_mixture_1} and \ref{alg_ln:update_mixture_2}, while consolidation of their weights are according to lines \ref{alg_ln:update_mix_w_1} and \ref{alg_ln:update_mix_w_2}. The merging of two weighted Gaussians into a single Gaussian approximation is as suggested in \cite{singleGaussianofGMM}. Note that the merging process is within a mode and if multiple modes are predicted at the same time a multimodal state distribution (GMM) is formed.

\begin{figure}[t]
\hrule
\vspace{0.2em}
\textbf{Algorithm 2} Long-term pred.~with~probabilistic~switching\label{alg:long_term_pred}
  \vspace{0.2em}
  \hrule
\begin{algorithmic}[1]
\vspace{0.2em}
\REQUIRE Initial state distribution: $p(\vxi[0])$, Policy: $\policyprob$
\STATE Init a segment $s$ with $p(\vxi[0]^s)=p(\vxi[0])$ and weight $w_0^s=1$
\FOR{$t=0$ to $T$}
    \FOR{each $s$\ s.t.\ $w_t^s>0$} \label{alg_ln:segment_loop}
        \STATE Obtain $p(\vxi[t]^s, \vui[t]^s), p(i_{t+1}^s)$ \qquad(\ref{eqn_ln:pol_pred_1_step}), (\ref{eqn_ln:p_xu_1_step}), (\ref{eqn_ln:mode_pred_1_step})
        \FOR{each $p(i_{t+1}^s=i') \ne 0 $} \label{alg_ln:mode_loop}
            \STATE Obtain $p(\vxi[t]^{s|s'}, \vui[t]^{s|s'})$ \label{alg_ln:split}
            \IF{$s=s'$}\label{alg_ln:prop_cont}
                \STATE Get $p(\vxi[t+1]^{s|s'})$ from $p(\vxi[t]^{s|s'}, \vui[t]^{s|s'})$ \label{alg_ln:mode_gp}\qquad(\ref{eqn_ln:nosw_1_step})
                \STATE Merge $p(\vxi[t+1]^{s|s'})$ to $p(\vxi[t+1]^{s})$\label{alg_ln:update_mixture_1}
                \STATE $w_{t+1}^{s} \mathrel{+}= p(i_{t+1}^s=i')w_t^s$\label{alg_ln:update_mix_w_1}
            \ENDIF
            \IF{$s\ne s'$} \label{alg_ln:prop_switch} 
            	\STATE Init $s'$ if required
                \STATE Get $p(\vxi[t+1]^{s|s'})$ from $p(\vxi[t]^{s|s'}, \vui[t]^{s|s'})$\qquad(\ref{eqn_ln:sw_1_step}) \label{alg_ln:init_gp}
                \STATE Merge $p(\vxi[t+1]^{s|s'})$ to $p(\vxi[t+1]^{s'})$ \label{alg_ln:update_mixture_2}
                \STATE $w_{t+1}^{s'} \mathrel{+}= p(i_{t+1}^s=i')w_t^s$\label{alg_ln:update_mix_w_2}
            \ENDIF
        \ENDFOR
    \ENDFOR
\ENDFOR
\vspace{0.2em}
\hrule
\end{algorithmic}
\end{figure}

\section{Experimental Results}
We conducted two experiments to validate our method. First, a simulated block of mass (a 1D simplification of a manipulator) was controlled to slide on a surface (1D) subject to abrupt changes in dynamics. Here, we also demonstrate the multimodal state propagation aspect. Second, a 7-DOF robot arm (YuMi) was controlled to move such that it experienced unexpected contacts and sliding motions. In this case, our method was applied to a dynamics model with $\vxi\in\mathbb{R}^{14}$ and $\vui\in\mathbb{R}^{7}$. For both experiments, we report results with small and large datasets. This helps in analyzing how the methods scale with data size. Each dataset is split into train ($\mathbb{D}^{train}$) and test ($\mathbb{D}^{test}$) subsets. We measure the performance of long-term predictions as average negative log likelihood (NLL) and average root mean square error (RMSE) of all time step predictions w.r.t the test trajectories. The RMSE is calculated for the most likely mode. One-step prediction performance is left out because long-term prediction is what matters for MBRL and also the latter subsumes the former.

As baselines, we used GP (SE kernel with ARD), manifold GP (mGP) \cite{calandra2016manifold} and the uncertainty-aware ANN method in PETS \cite{chua2018deep} (uANN). In mGP, the parameters of an ANN feature space mapping and an SE kernel based on that feature space is jointly learned. The extra ANN based mapping was designed to provide enough flexibility to represent discontinuities. The uANN model is an ensemble of many ANN models (bootstraps) and the variance in their predictions is interpreted as uncertainty of the overall model. It benefits from the high flexibility and scalability of ANNs. For all the three baselines, to generate probabilistic long-term prediction, we propagated 50 particles up to the full episode lengths and followed it by density estimation (DPGMM) at each time step.

 
For DPGMM, we used the implementation in \cite{scikit-learn} which requires an upper bound on the number of clusters $K_{max}$ to be specified. We used the values 10 and 20 for the sliding mass and robot experiments, respectively. The concentration parameter $\alpha$, known to be generally robust \cite{escobar1995bayesian}, was set to $0.1$. For the UT method, we used the recommended values \cite{julier1997new} for its hyperparameters. For both GP and mGP, we used the standard practice of marginal likelihood optimization to determine hyperparameters. The SVM model was trained using an automated grid search. For uANN we chose 5 ensembles for all cases and the TS$\infty$ uncertainty propagation scheme was used (when possible), in which particles are assigned to one of the ensembles permanently \cite{chua2018deep}. All ANN network structures were fully connected layers. Although our approach utilized a number of machine learning methods, in almost no cases did we encounter difficult hyperparameter tuning.

\begin{figure}[t]
\centering
  \includegraphics[width=0.45\textwidth]{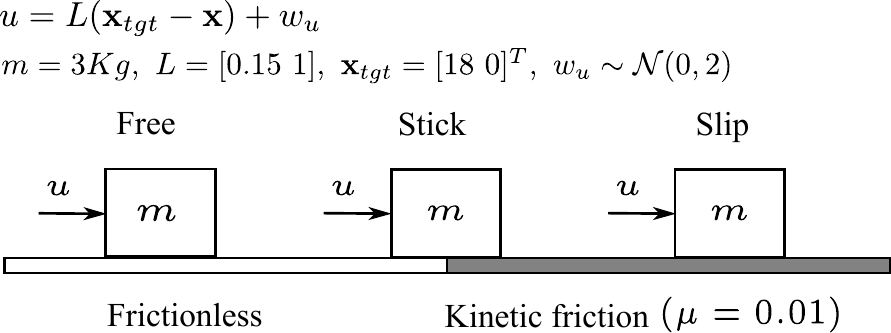}
  \caption{Three scenarios of the sliding mass experiment}
  \label{fig:sliding_mass}
\end{figure}

\subsection{The Sliding Mass Experiment}\label{sec:block_exp}
A block of mass sliding on a surface, controlled by a policy (linear Gaussian regulator) that pulls in one direction, is subjected to three dynamics modes: free frictionless motion, halted due to sticking and slipping under kinetic friction (Fig. \ref{fig:sliding_mass}). After experiencing the first mode it gets blocked at a point of high static friction which the stochastic control policy may or may not overcome, thus resulting in a probabilistic scenario of either staying stuck or slipping. We expect discontinuities at stick and also slip. The slip is implemented as an instantaneous velocity jump to 5 m/s. 

\begin{figure}[t]
     \centering
    \subfloat{\includegraphics[width=0.49\textwidth]{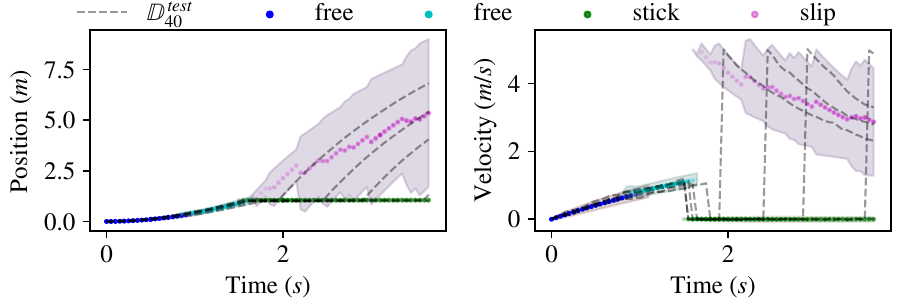}}\\
    \vspace{-0.5em}
     \subfloat{\includegraphics[width=0.49\textwidth]{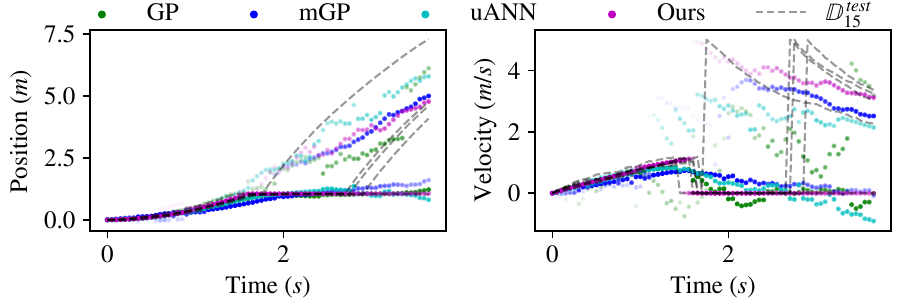}}
    \caption{Long-term prediction of sliding mass. Top \textbf{(\ref{fig:lt-pred_blocks}a)}: Prediction of our method ($\mathbb{D}_{40}$). Bottom \textbf{(\ref{fig:lt-pred_blocks}b)}: Comparison of the predicted mean with baselines ($\mathbb{D}_{15}$). The transparency of mean points is according to their probabilities. The $2\sigma$ regions for individual modes are not adjusted for their weights.}
    \label{fig:lt-pred_blocks}
\end{figure}

The experiment generated two datasets: $\mathbb{D}_{15}=\{\mathbb{D}_{15}^{train}, \mathbb{D}_{15}^{test}\}$ consisting of $15+5$ trials, and $\mathbb{D}_{40}=\{\mathbb{D}_{40}^{train}, \mathbb{D}_{40}^{test}\}$ consisting of $40+5$ trials. Each trial trajectory is simulated for $T=75$ time steps with a stepping time of $0.05$ seconds. Fig. \ref{fig:lt-pred_blocks}a shows the results with $\mathbb{D}_{40}$. We used the same policy and initial state distribution that were used for generating the data. The prediction closely follows the test trajectories ($\mathbb{D}_{40}^{test}$) while handling the two discontinuities well. Multimodality during switching can be seen as segment overlaps. Multimodal propagation of slip and stick is also evident. A switch without any discontinuity appears in the middle of the free motion indicating unnecessary mode identification. Such over-identification is practically advantageous since it reduces the GP training time. The variance prediction of our model is generally consistent (test data inside the $2\sigma$ region) but with slight violations.

\begin{figure}[t]
\centering
  \includegraphics[width=0.3\textwidth]{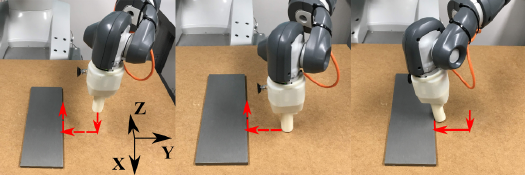}
  \caption{Setup for the contact motion experiment. The task environment is rigid and static. From left to right: Free motion seeking contact, sliding motion constrained by $X-Y$, and sliding motion constrained by $X-Y$ and $Z-X$.}
  \label{fig:yumi_exp}
\end{figure}

\begin{table}
 \caption{Sliding mass scores (avg. 10 restarts)}
 \label{tab:sliding_mass}
\centering
\begin{tabular}{ |c|c|c|c|c|c| } 
 \hline
 & \multicolumn{2}{c|}{$\mathbb{D}_{15}$} & \multicolumn{2}{c|}{$\mathbb{D}_{40}$} \\ 
  \hline
   & NLL & RMSE & NLL & RMSE \\ 
 \hline
 GP & $0.84\pm3.04$ & $1.77$ & - & - \\ 
 mGP & $1.8\pm2.94$ &  $3.0$ & - &  -\\ 
 uANN & $0.92\pm3.57$ & $2.75$ & $0.57\pm3.3$ & $3.53$\\ 
 \textbf{Ours} & $\textbf{-2.32}\pm\textbf{3.8}$ & $\textbf{0.63}$ & $\textbf{-2.3}\pm\textbf{3.45}$ & $\textbf{0.76}$ \\ 
 \hline
\end{tabular}
\end{table}

We used network sizes of [32, 32, 2] and  [16, 16, 16, 2] for mGP and uANN, respectively. Figure \ref{fig:lt-pred_blocks}b and Table \ref{tab:sliding_mass} show comparisons of the various methods. The plot shows only mean values, but both the means and variances are used for the score calculation. Our method accurately handled both discontinuities. None of the baselines were successful in handling the free-to-stick discontinuity, while both uANN and mGP partially succeeded in the other. Although the GP model failed in both cases, when averaged over $10$ restarts, it gave a better score than uANN and mGP. Our method has the best scores in all cases. uANN showed a slight advantage over mGP with $\mathbb{D}_{15}$. GP and mGP did not scale to $\mathbb{D}_{40}$ (long training times). $\mathbb{D}_{40}$ did not lead to better scores although some slight reduction in the spread of NLL is noticeable. Our method occasionally underestimated the variances while all the baselines consistently overestimated them. 

\subsection{The Contact Motion Experiment}\label{sec:robot_exp}
The experimental setup consisted of the bimanual 7-DOF robot YuMi mounted on a wooden platform ($X-Y$ plane), which also had a thick steel plate rigidly attached on its surface (Fig. \ref{fig:yumi_exp}). The robot interacts with the environment through a rigidly attached peg at the end-effector. A Cartesian space trajectory (0.06 m/s) was generated such that it caused: free motion along $Z$-axis, contact with the platform, sliding motion along $Y$-axis, contact and sliding motion along the edge of the steel plate ($X$-axis). Stable contacts were ensured by designing the reference trajectories beyond physical constraints and setting low gains for the joint space proportional-derivative (PD) controllers. A closed-loop inverse kinematics scheme was also employed. Gaussian noise ($\mathcal{N}(0,10^{-4})$) was added to the $X$ and $Y$ components of the reference trajectory as exploration noise. This experiment was carefully designed to feature the main elements of contact-rich motion: discontinuous change in velocity, sliding on different materials, and motion constrained by one or more surfaces. Note that our method can work with any policy. The engineered impedance controller is only for experimental purposes.  

A trial consisted of 100 time steps sampled at 0.05 seconds interval. We also perturbed the base controller by altering the joint space proportional gains of all axes by $\pm2\%$, $\pm5\%$, and $\pm10\%$. The experiment generated two datasets: $\mathbb{D}_{15}=\{\mathbb{D}_{15}^{train}, \mathbb{D}_{15}^{test}\}$ and $\mathbb{D}_{40}=\{\mathbb{D}_{40}^{train}, \mathbb{D}_{40}^{test}\}$, where $\mathbb{D}_{15}^{train}$ contained 15 trials from the base policy and $\mathbb{D}_{40}^{train}$ had extra 5 trials each from all except the $-10\%$ case. The $-10\%$ case also had 5 trials and was designated as the test set $\mathbb{D}_{40}^{test}$ ($=\mathbb{D}_{15}^{test}$). By perturbing the controller (policy), we aim to closely emulate a typical MBRL iteration in which the dataset for model learning would be a mixture of trials from slightly different policies. Although the controller is a complex Cartesian space controller, we extracted the generated joint space reference trajectory and PD gains of the test case and formed a joint space equivalent to be used as the test policy for long-term prediction.

\begin{figure}[t]
     \centering
     \vspace{-1em}
    \subfloat{\includegraphics[width=0.245\textwidth]{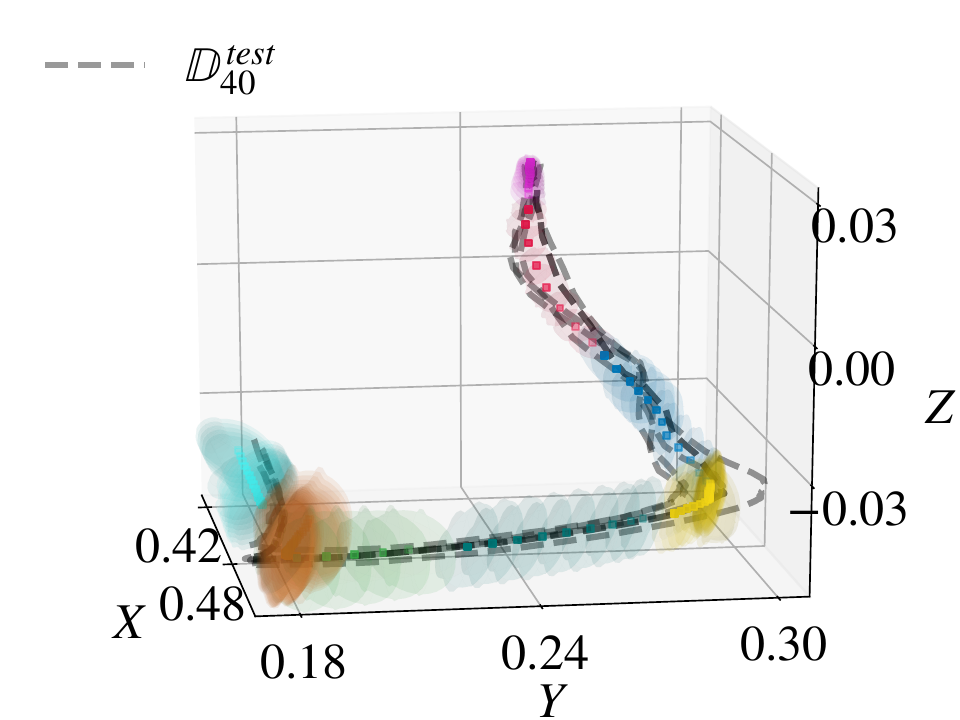}}%
     \subfloat{\includegraphics[width=0.245\textwidth]{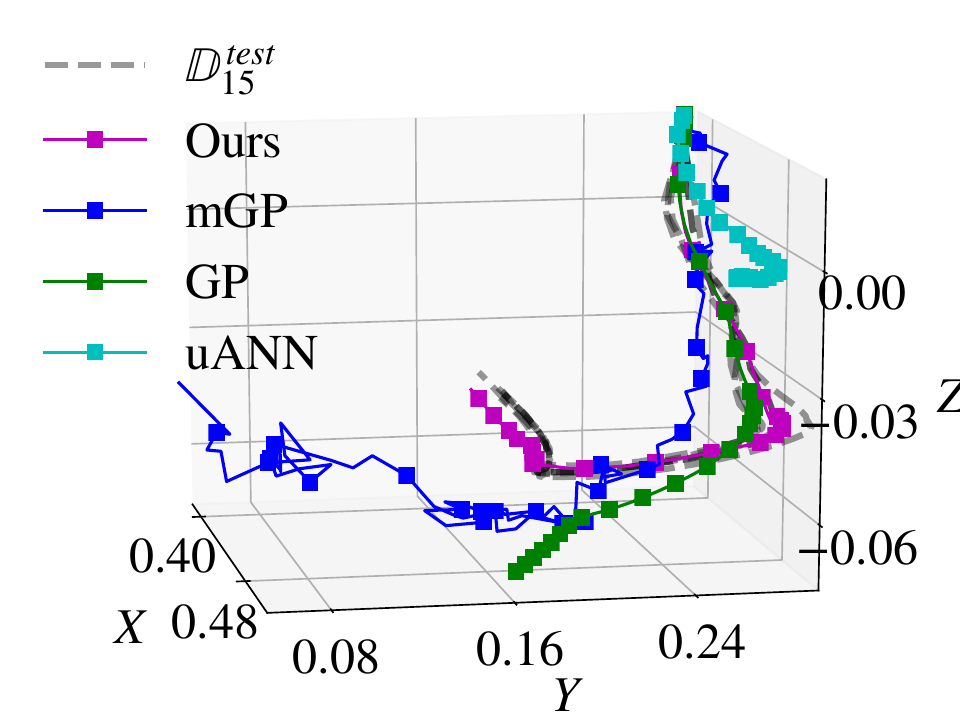}}\\
     \vspace{-.5em}
      \subfloat{\includegraphics[width=0.485\textwidth]{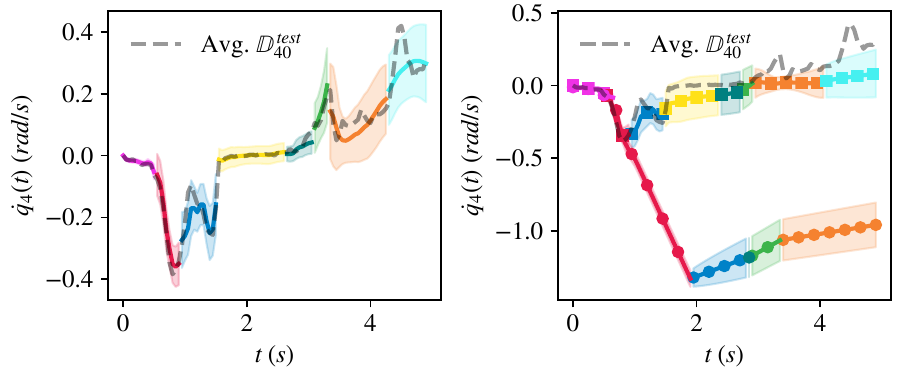}}
      \vspace{-1.em}
    \caption{Long-term prediction of contact motion. Top-left \textbf{(\ref{fig:contact_pred}a)}: Prediction of our method ($\mathbb{D}_{40}$). Top-right \textbf{(\ref{fig:contact_pred}b)}: Comparison with baselines ($\mathbb{D}_{15}$). Bottom-left \textbf{(\ref{fig:contact_pred}c)}: Prediction of joint 4 velocity ($\dot{q}_4(t)$). Bottom-right \textbf{(\ref{fig:contact_pred}d)}: $\dot{q}_4(t)$ prediction with switching GP model (circle) and our method minus reset map (square). Colors correspond to modes in \ref{fig:contact_pred}a, \ref{fig:contact_pred}c \& \ref{fig:contact_pred}d.}
     \label{fig:contact_pred}
\end{figure}

In Fig. \ref{fig:contact_pred}a, the results of $\mathbb{D}_{40}$ are presented in the Cartesian space (only translational position) for visual clarity. The transformation from the joint space was done by applying the UT method on forward kinematics. The plot shows only the most likely mode at each time step. It can be seen that the method sequentially switched through eight modes (different colors) indicating over-clustering. The prediction manages to closely follow the ground truth trajectories, respecting the two instances of contacts (free-to-wood and wood-to-steel), and with fairly consistent variance predictions (2$\sigma$). We take the opportunity to compare some common alternatives. In Fig. \ref{fig:contact_pred}d we see that a simple GP based switching model \cite{calandra2015learninga} cannot achieve long-term prediction at all, while our method minus the reset map cannot handle discontinuities (at 1.5 and 3.5 seconds) in state propagation. Our complete method (Fig. \ref{fig:contact_pred}c) is able to achieve effective switching and can handle discontinuities.

The performance of our method is significantly better than the baselines, none of which succeeded in closely following the ground truth (Fig. \ref{fig:contact_pred}b). We used network sizes of [32, 32, 3] and [64, 64, 64, 14] for mGP and uANN, respectively. While attempting the TS$\infty$ propagation scheme, it was revealed that although it performed generally well up to a relatively shorter horizon, as was used in \cite{chua2018deep}, it ultimately went unstable and predicted large values. We reduced this problem by averaging predictions of all ensembles for each particle (different from TS$\infty$). Larger training data also alleviated this problem. While our method predicted fairly consistent variances, with occasional underestimations, all of the baselines produced overestimated variances. Our method has the best scores in all cases (Table \ref{tab:yumi_exp}). mGP has better scores than uANN but the plot shows noisy mean predictions. As in the previous experiment, the GP outperformed the other baselines. Both GP and mGP did not scale to $\mathbb{D}_{40}$. Our method showed slight improvement with $\mathbb{D}_{40}$, but the case with uANN was inconclusive. For $\mathbb{D}_{40}$, the training times were 53s and 923s for uANN and our method, respectively.

\begin{table}[t]
\centering
\caption{Contact motion scores (avg. 10 restarts)}\label{tab:yumi_exp}
\begin{tabular}{ |c|c|c|c|c|c| } 
 \hline
  & \multicolumn{2}{c|}{$\mathbb{D}_{15}$} & \multicolumn{2}{c|}{$\mathbb{D}_{40}$} \\ 
  \hline
   & NLL & RMSE & NLL & RMSE \\ 
 \hline
 GP & $-4.6\pm 38$ & $0.25$ & - & - \\ 
 mGP & $-0.9\pm12$ &  $1.22$ & - &  - \\ 
 uANN & $5.1\pm23$ & $2$ & $71.4\pm82$ & $0.47$ \\ 
 \textbf{Ours} & $\textbf{-22.7}\pm\textbf{26}$ & $\textbf{0.17}$ & $\textbf{-24}\pm\textbf{27}$ & $\textbf{0.15}$\\ 
 \hline
\end{tabular}
\end{table}

\section{Discussions}\label{sec:disscuss}
The results validate that a hybrid automata structure is advantageous for model learning and long-term prediction of contact-rich manipulation tasks in low data regime. More specifically, specialized local GP models combined with the concept of reset maps is what contributed the most. The chosen baselines (uANN and mGP) are both highly capable methods which, if given enough data, could potentially learn internal representations that match a hybrid structure; but, such amount of data is impractical in a real-world MBRL. Furthermore, given that contact dynamics are notoriously hard to model, simulation to real transfer is also a challenge for contact-rich tasks. This leaves methods such as ours as a promising choice. However, our method may not scale as well as the uANN method. This is because a GP based approach, even with a divide and conquer strategy such as ours, may still face difficulty in scaling with the training data as MBRL progresses. Straightforward solutions include parallelized implementation of GP training (each output dimension of every mode is an independent GP) and restricting the number of previous iterations to include in model learning within MBRL. 


The proposed method relied on clustering to discover dynamics modes. The kinematic feature in the operational space is informative for encoding rigid contact events and also ensured linear separability (for clustering) that significantly simplified our model learning method. Note that the identified clusters are not always uniquely corresponded to actual modes. This is because not all mode transitions are accompanied by impacting contacts and the i.i.d assumption makes DPGMM easily identify multiple clusters within a mode. However, this is not concerning because our focus is modeling and prediction involving discontinuities rather than revealing the exact modes. The clustering strategy could be affected by very sparse samples but can be improved by segmenting trajectories with considerations to temporal correlation in data~\cite{Yin2018-ID1003}. 


A number of points are worth discussing in the context of MBRL. Although our method is strictly limited to model learning, any policy that may be produced in an MBRL algorithm, can be used with it. The policy induced state-action distribution mismatch between prediction time and training time is a common issue for MBRL algorithms. Our GP-based model features an inherent ability to encode uncertainty due to lack of data as demonstrated in \cite{deisenroth2015gaussian}. Still, it is the design of policy search strategy that eventually regulates the policy shift in order to minimize this mismatch. Hence, any further consideration is outside the scope of model learning. If an unseen mode is encountered during rollouts, the potentially different dynamics could set back a particular iteration. But, the data collected will reveal the new mode in the next iteration and the MBRL algorithm will continue to make progress. An MPC approach to MBRL \cite{chua2018deep} \cite{nagabandi2018neural} is conceivable but the evaluation of online performance is not included since we only consider a policy search setting. Finally, our particle based method does not offer the possibility to compute analytic gradients, thereby limiting it only for gradient-free approaches such as \cite{chatzilygeroudis2017black}.

\section{Conclusion}
In this paper, we addressed the problem of predicting the state evolution of contact-rich manipulation tasks using learned dynamics models. In MBRL, such predictions are useful for evaluating a policy. Focusing on discontinuities and data efficiency as the main challenges, we show that by adopting the structure of hybrid automata and GP as the main function approximator significant improvement in data efficiency and prediction accuracy is attainable. The proposed method was tested on conditions resembling a typical MBRL iteration and compared with highly flexible baselines. The need for achieving further scaling with data remains and a clear strategy has been suggested. Therefore, although limited to gradient-free settings, the proposed method is a promising option for model learning for contact-rich tasks.

\bibliographystyle{IEEEtran}
\bibliography{References/rl,References/rl_skill,References/rl_skill_compliant,References/other,References/imitation_learning,References/control_opt_robotics,References/model_learning,References/ml,References/IEEEabrv,References/for_this_doc}
\end{document}